\documentclass[conference]{IEEEtran}
\IEEEoverridecommandlockouts

\usepackage[OT1]{fontenc} 

\usepackage{cite}
\usepackage{amsmath,amssymb,amsfonts}
\usepackage{algorithm}
\usepackage{algorithmic}
\usepackage{graphicx}
\usepackage{textcomp}
\usepackage{xcolor}
\usepackage{booktabs}
\usepackage{tabularx}
\usepackage{pgf-pie}
\usepackage{pgfplots}
\pgfplotsset{compat=1.18}
\usepackage{url}

\pdfoutput=1

\begin{document}

\title{Few-Shot Prompting for Extractive Quranic QA with Instruction-Tuned LLMs}

\author{
\IEEEauthorblockN{Mohamed Basem}
\IEEEauthorblockA{
\textit{Faculty of Computer Science} \\
\textit{MSA University}\\
Giza, Egypt \\
mohamed.basem1@msa.edu.eg}
\and
\IEEEauthorblockN{Islam Oshallah}
\IEEEauthorblockA{
\textit{Faculty of Computer Science} \\
\textit{MSA University}\\
Giza, Egypt \\
 islam.abdulhakeem@msa.edu.eg}
\and
\IEEEauthorblockN{Ali Hamdi}
\IEEEauthorblockA{
\textit{Faculty of Computer Science } \\
\textit{MSA University}\\
Egypt \\
ahamdi@msa.edu.eg
}
\and
\IEEEauthorblockN{Ammar Mohammed}
\IEEEauthorblockA{
\textit{Faculty of Computer Science} \\
\textit{MSA University}\\
Egypt \\
ammohammed@msa.edu.eg
}
}
% =========================================
\IEEEoverridecommandlockouts            % allow \IEEEpubid
\IEEEpubid{\makebox[\columnwidth]{979-8-3315-0185-3/25/\$31.00 ©2025 IEEE\hfill}
\hspace{\columnsep}\makebox[\columnwidth]{}}  % balances the columns
\maketitle                              % create the title
\IEEEpubidadjcol                        % push text below the notice
% =========================================

\begin{abstract}
This paper presents two effective approaches for Extractive Question Answering (QA) on the Qur’an. It addresses challenges related to complex language, unique terminology, and deep meaning in the text. The second uses few-shot prompting with instruction-tuned large language models such as Gemini and DeepSeek. A specialized Arabic prompt framework is developed for span extraction. A strong post-processing system integrates subword alignment, overlap suppression, and semantic filtering. This improves precision and reduces hallucinations. Evaluations show that large language models with Arabic instructions outperform traditional fine-tuned models. The best configuration achieves a pAP@10 score of 0.637. The results confirm that prompt-based instruction tuning is effective for low-resource, semantically rich QA tasks.
\end{abstract}

\vspace{3mm}

\begin{IEEEkeywords}
Quranic Question Answering, Machine Reading Comprehension, Classical Arabic, Few-Shot Prompting, Large Language Models, Instruction-Tuning, Fine-Tuning, Arabic NLP, Span Extraction.
\end{IEEEkeywords}
\section{Introduction}
Recent advances in large language models (LLMs) have significantly transformed the field of machine reading comprehension (MRC). These advances allow systems to interpret, analyze, and answer more semantically specific questions~\cite{kazi2023survey}. These models have shown particular promise in areas requiring a deeper understanding of context and language.

Religious texts present a unique and underexplored challenge, particularly \textbf{Quranic Question Answering (QA)}, which requires not only linguistic competence~\cite{alrayzah2023challenges}. The Qur'an is written in classical Arabic. This language is known for its complicated morphology, flexible syntax, and symbolic expressions~\cite{salama2018qur}. These features make automatic span extraction especially difficult. The model must determine between literal and metaphorical languages. It must handle domain-specific vocabulary. It also needs to identify implied connections in verses with nonlinear structures~\cite{malhas-etal-2023-quran,oshallah2025cross}.

Moreover, questions posed to the Qur'an differ from general MRC tasks. General MRC questions are often based on facts or events. Quranic questions, however, often look for answers with spiritual, moral, or interpretive meaning. The absence of diacritical marks (tashkeel) complicates accurate retrieval. Ellipses, metaphors, and the linked themes across chapters make it harder for traditional QA systems to work well in this area~\cite{basem2024optimized,osmanhamoud2017question}.

To address these challenges, the \textbf{Qur’an QA 2023 Shared Task} introduced a benchmark. This benchmark is specifically adapted for extracting MRC on Qur’anic passages~\cite{malhas-etal-2023-quran}. Includes accurate annotations. It also features span-based evaluation metrics. In addition, it provides a standardized task setup (TaskB). This setup focuses on identifying exact subphrases in paragraphs that answer specific questions. The benchmark offers a crucial platform for evaluating QA systems. It is particularly relevant for religious and low-resource contexts.

We investigate two complementary modeling strategies. (1) involves fine-tuning Arabic transformer-based models such as AraBERT and AraELECTRA on task-specific data. (2) focuses on few-shot prompting of instruction-tuned LLMs like Gemini and DeepSeek. This uses structured Arabic templates. By comparing these approaches, we aim to assess their relative advantages. We evaluate full parameter fine-tuning against instruction-based prompting in a zero/few-shot setting.

To further improve the performance of the system, we have developed post-processing pipelines that address common issues in span extraction. The pipelines include correcting the alignment of subwords. They also involve resolving non-maximum suppression of overlapping limits. In addition, they remove vague or redundant answers~\cite{zhang2021comparing}. These improvements adapt to the subtleties of classical Arabic. They also meet the requirements of extractive QA.

The contributions of this paper are presented in a single sentence: We design specialized fine-tuning and prompting strategies that account for the linguistic and thematic characteristics of Quranic Arabic, implement a post-processing module that refines raw model predictions using token-level filtering, semantic validation, and re-ranking, and conduct an extensive evaluation using multiple Quranic QA datasets to demonstrate state-of-the-art performance and highlight the strengths of instruction-tuned LLMs in low-resource, high-context domains.

The results show that carefully designed prompts help LLMs adapt to instructions. These prompts achieve better performance than transformer-based models. This is true even in tasks requiring high span accuracy and religious sensitivity. The study showed that prompt engineering is an effective alternative to fine-tuning QA extraction.

\section{Related Work}
The Qur'anic language challenges answer-extraction systems with complex syntax and multiple-meaning words.
It has unique linguistic features related to its expressive nature~\cite{essam2024decoding}. The Qur'an QA 2023 Shared Task (Task B) targeted answer span extraction. It used Qur'anic passages for answers. It tackled linguistic complexity issues. It addressed semantic richness challenges. The task was in a low-resource domain where standard QA approaches often underperform~\cite{malhas-etal-2023-quran}.

Several teams addressed TaskB with innovative approaches to tackle Qur'anic Arabic’s difficulties. Two teams, Tce~\cite{elkomy2023tce} and LKAU23~\cite{alnefaie2023lkau23}, used similar methods. They employed transformer-based ensembles. They fine-tuned on QRCD v1.2 dataset. This adapted to Qur'anic text characteristics. Ensembling improved prediction consistency. Post-processing refined answer spans. This aligned predictions with passages. It targeted span boundary detection. This is key for extractive QA. It showed the effectiveness of ensembling, particularly in low-resource settings where domain adaptation is well suited.

They differed in their specific approaches. For Tce, the approach used a transformer-based ensemble. It had a multi-answer loss function. This handled multiple answer spans. A thresholding mechanism detected unanswerable questions. The model used a transformer-based ensemble with multi-answer loss, fine-tuned on QRCD v1.2, with post-processing for span refinement. Tce achieved a pAP@10 of 0.571. They took first rank in this shared task for TaskB.

For LKAU23, the approach used AraBERT models. These were AraBERT-base and AraBERT-large. They included Tashkeela, OSIAN, ARCD, and QUQA datasets. This improved Arabic coverage. They focused on ensemble predictions and lacked a specialized loss function. The models used were an AraBERT-base ensemble and an AraBERT-large ensemble, fine-tuned on QRCD v1.2 and augmented datasets. They achieved a pAP@10 of 0.49.

The Al-Jawaab team used a two-stage system for extractive question answering. They applied OpenAI’s text-embedding-ada-002 for dense passage retrieval. This identified relevant Qur'anic passages. GPT-3.5 generated unconstrained semantic responses. These captured broad contextual understanding. GPT-4 extracted precise answer spans from retrieved passages. This ensured accurate span boundary detection. The model used was a hybrid system combining text-embedding-ada-002, GPT-3.5, and GPT-4. The hybrid approach balanced semantic flexibility and extractive accuracy. It effectively addressed Qur'anic Arabic’s linguistic challenges. They achieved a pAP@10 of 0.539~\cite{zekiye2023aljawaab}. They took second rank.

The GYM team adopted a multi-task transfer learning approach. They used an AraElectra-based architecture. They fine-tuned for answer span detection. This leveraged domain-specific adaptation. The model used was an AraElectra-based multi-task model, fine-tuned on QRCD v1.2. The approach improved performance in low-resource settings. It focused on precise span boundary modeling. This addressed Qur'anic text complexities. They achieved a pAP@10 of 0.461~\cite{mahmoudi2023gym}. This improved 0.23 over the baseline.

These efforts highlight key trends. Accurate span detection is crucial. Domain adaptation aids low-resource settings. Tailored QA architectures address complexities. Customized MRC models are needed for religious texts.

\section{Methodology}

\subsection{Task Definition}

The problem addressed by Shared Task B of the Qur'an Question Answering focuses on 
\textbf{MRC}. Given a passage of the Qur'an $P$ and a natural language question $Q$, the goal is to extract from $P$ one or more spans of text $A1, A2,..., An$ that best answer $Q$. The answer must be an \textbf{exact subspan} from the passage, not a generated answer or a paraphrased answer. This requires systems to identify the boundaries of answer spans properly, resting on an appropriate semantic and contextual comprehension of the question and passage~\cite{malhas-etal-2023-quran}.

\begin{figure}
    \centering
    \includegraphics[width=1\linewidth]{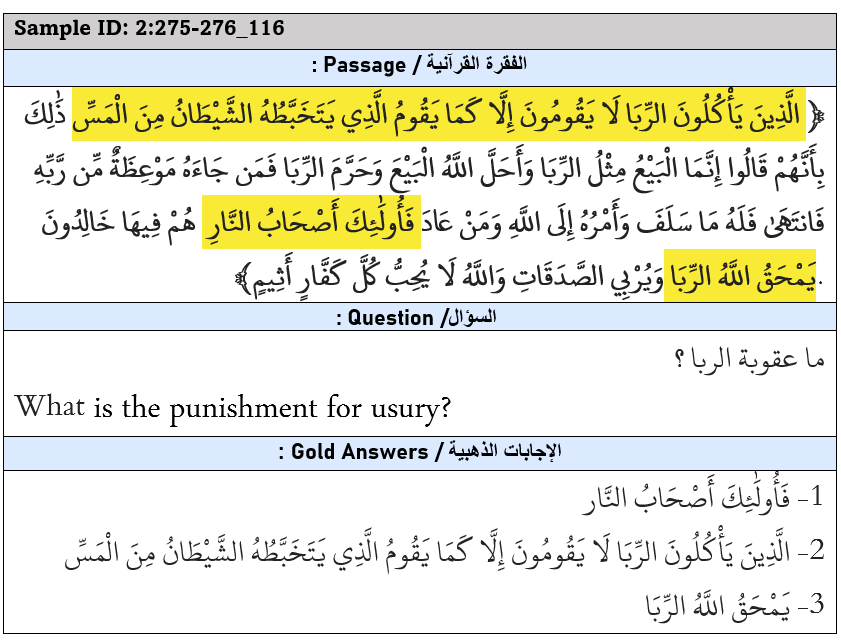}
    \caption{An example of the MRC task: Highlighted are the exact Quranic phrases identified by the model as answering the question about the punishment for usury, showcasing its ability to extract precise spans from complex religious texts.}
    \label{fig:example}
\end{figure}

Figure~\ref{fig:example} The task is illustrated in Figure 1. The example highlights a question and multiple answer spans extracted from the passage.. It underscores the importance of precise span detection and semantic comprehension in Quranic MRC.

Formally, the task is defined as:
\begin{itemize}
    \item \textbf{Input:} A question $Q$ in Modern Standard Arabic and a passage $P$ from the Qur’an.
    \item \textbf{Output:} A set of one or more exact answer spans $\{A_1, A_2, ..., A_n\} \subset P$.
\end{itemize}

\subsection{Dataset Preparation}
To support strong model training and evaluation, we used three main datasets: the Quranic Understanding Question Answering (QUQA), the Arabic Reading Comprehension Dataset (ARCD), and the Qur'anic Reading Comprehension Dataset version 1.2 (QRCDv1.2). This last dataset was released as part of the Quran QA 2023 Shared Task. Each dataset plays a unique role in improving the model's generalization for different MRC tasks in Arabic and Quranic contexts.

The \textbf{QUQA dataset} \cite{quqa2023}consists of selected question-passage pairs from the Quran, emphasizing a wide range of topics and language styles. It provides high-quality annotations suitable for supervised learning.

The \textbf{ARCD dataset} \cite{arcd2021} is a benchmark for Arabic reading comprehension, containing modern standard Arabic (MSA) passages and questions. Although originally constructed for general mrc tasks. It enables the model to understand Arabic linguistic structures and improve its performance on Quranic texts.

The \textbf{QRCDv1.2 dataset} \cite{malhas-etal-2023-quran} provides domain-specific annotated data focused on passage retrieval and reading comprehension tasks for Quranic texts. It served as the primary resource for fine-tuning and evaluating models in these experiments. The distribution of question-passage (QP) pairs and question-passage-answer (QPA) triplets across training, development, and test sets is shown in Table~\ref{tab:qrcd_distribution}.

\begin{table}[htbp]
\caption{Distribution of QP pairs and QPA triplets in QRCDv1.2. QP refers to a question and its passage; QPA includes an answer span.}
\centering
\scriptsize
\setlength{\tabcolsep}{1pt}
\begin{tabularx}{\columnwidth}{lcccc|c}
\toprule
\textbf{Type} & \textbf{Train} & \textbf{Dev} & \textbf{Test} & \textbf{All} & \textbf{QPA} \\
\midrule
Multi-answer & 134 (14\%) & 29 (18\%) & 62 (15\%) & 224 (14\%) & 552 (29\%) \\
Single-answer & 806 (81\%) & 124 (76\%) & 331 (81\%) & 1,261 (81\%) & 1,261 (67\%) \\
Zero-answer & 52 (5\%) & 10 (6\%) & 14 (4\%) & 76 (5\%) & 76 (4\%) \\
\midrule
\textbf{Total} & \textbf{992} & \textbf{163} & \textbf{407} & \textbf{1,562} & \textbf{1,889} \\
\bottomrule
\end{tabularx}
\label{tab:qrcd_distribution}
\end{table}

We adapted external datasets by reformatting them to align with the structure used in the Quran QA 2023 Shared Task (QRCDv1.2), ensuring consistency in passages, questions, and answer annotations.

For API-based models like Gemini and DeepSeek, we constructed three-shot prompts using examples from QRCDv1.2. For transformer-based models trained locally, we combined all datasets into a unified training corpus following the standardized format.

This reformatting step enabled consistent model training and evaluation across diverse Arabic and Quranic QA sources.

\subsection{Model Architecture}

This system integrates two model types: API-accessed LLMs and fine-tuned transformer-based models.

We adopt a three-shot prompting approach for LLMs such as Gemini and DeepSeek. Each prompt includes a Quranic passage, a question, and a sample answer span, allowing the model to infer and generate span-level predictions. These models are accessed through an API and can be used in a zero-shot or few-shot setting without changing the parameters. 

Transformer-based models are built with the SimpleTransformers library and initialized from checkpoints that were pre-trained on TyDi QA.We include Arabic-specific architectures like AraELECTRA, CAMeLBERT, and AraBERT. These models are fine-tuned on a single set of datasets that includes QUQA, ARCD, and QRCDv1.2. The fine-tuned models use a span extraction framework that consists of:
\begin{itemize}
    \item \textbf{Input Representation:} Tokenization and combining the passage and question with [SEP] tokens. 
    \item \textbf{Encoding Layers:}  Multi-layer transformers use self-attention to capture contextual relationships. 
    \item \textbf{Answer Head:}  A classification head predicts where the answer starts and ends within the passage.  
\end{itemize}

By combining prompt-guided LLMs and supervised fine-tuned models, the architecture enables both generative and extractive reasoning for Quranic QA. This improves flexibility across different tasks.

\subsection{Fine-Tuning Setup}
All models are initialized from pre-trained checkpoints trained on the TyDi QA dataset. The fine-tuning configuration is summarized in Table~\ref{tab:finetuning_setup}.

\begin{table}[ht]
\caption{Fine-Tuning Hyperparameters}
\label{tab:finetuning_setup}
\centering
\begin{tabular}{ll}
\hline
\textbf{Hyperparameter} & \textbf{Value} \\
\hline
Training batch size & 16 \\
Evaluation batch size & 16 \\
Learning rate & 2e-5 \\
Number of epochs & 10 \\
Evaluation frequency & Every 500 steps \\
Use multiprocessing & False \\
n-best size & 20 \\
Manual seed & 109 \\

\hline
\end{tabular}
\end{table}

Batch size 16 is selected for GPU memory utilization and gradient stability with BERT-type architectures. The learning rate of 2×10 is chosen due to its success in fine-tuning for question answering tasks and due to its stability so as not to overshoot the optimal minima.
The number of epochs was 10, with evaluation every 500 steps to keep track of overfitting and the learning +state. To make the procedure straightforward and reproducible, multiprocessing was deactivated with a fixed random number seed of 109. Only the best-performing checkpoint would be saved along training based on the evaluation performance-this allows for a solid, nuanced adaptation of Arabic pre-trained transformers for Quranic question answering.
\begin{figure*}[ht]
    \centering
    \includegraphics[width=\linewidth]{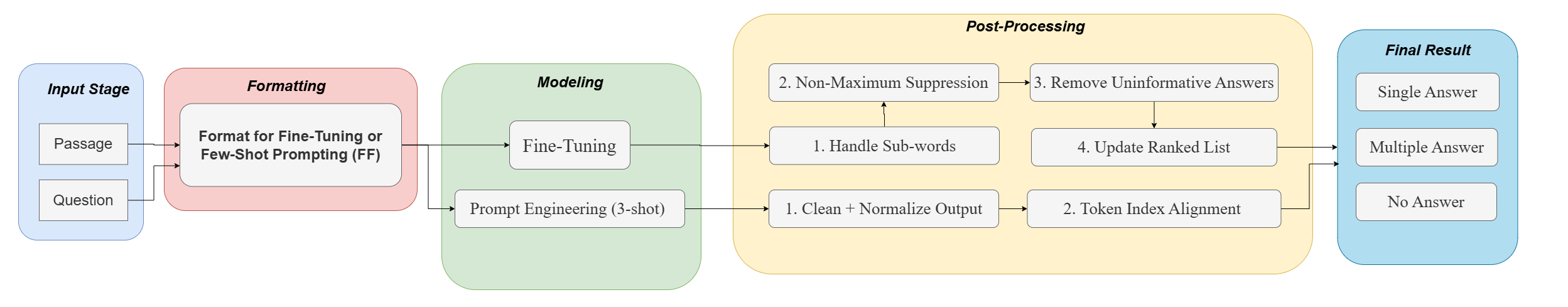}
    \caption{The Quran QA pipeline has five phases: (1) Input of a passage and question; (2) Formatting for fine-tuning or prompting; (3) Modeling using fine-tuned transformers or API-based LLMs with 3-shot prompts; (4) Post-processing for token alignment, normalization, and answer filtering; (5) Final output as one, multiple, or no answers.}
    
    \label{fig:enter-label}
\end{figure*}

\subsection{Few-Shot Learning}

The few-shot learning strategy was selected for the optimal performance enhancement of API-based LLMs like Gemini and DeepSeek. Such few-shot prompting requires a number of example instances to be provided in the prompt to better direct the models toward understanding the task requirements.

In our setup, three representative examples were selected from the QRCDv1.2. These examples included one multi-answer question, one single-answer question, and one zero-answer case. This diversity enabled the models to learn how to handle a range of answer scenarios, improving their robustness when dealing with Quranic comprehension tasks.

These few-shot demonstrations are seen as very important for guiding the models to extract the accurate spans from the Quranic text to reduce hallucinations while keeping a high fidelity to the actual text.

\subsection{Prompt Engineering}

Following this, we designed a structured Arabic prompt for API-accessed models like Gemini and DeepSeek to assist the model in generating API responses from Quranic passages. The prompt is carefully designed for precise and valid answer extraction. Here is the template we used:

\begin{itemize}
\item Task: Extracting Answers from Quranic Texts

\item Instructions:
\begin{itemize}
\item Extract all phrases from the passage that answer the question, whether the answer is direct or implied.
\item The answer must be a literal part from the passage.
\item If you cannot find a suitable answer, write: There is no answer in the given text.
\item Do not write any additional explanation or interpretation.
\item Try to understand the interpretation (tafsir) of each verse to reach an accurate answer.
\item Write the answers in the following format:
\begin{itemize}
\item ''First answer''
\item ''Second answer''
\end{itemize}
\item Examples: \{few\_shot\_text\}\
\item Answer the following passage:
\item Passage: \{passage\}
\item Question: \{question\}
\item Answers:
\end{itemize}
\end{itemize}

The few-shot learning technique adopted by us includes inserting three representative examples into the \texttt{\{few\_shot\_text\}} placeholder. We select examples from the Quran QA 2023 training that satisfy different criteria concerning answer types: one multi-answer question, one single-answer question, and one zero-answer case. Each example is accompanied by its Tafsir to ensure a deeper understanding of the verses. This diversity enables the language model to effectively learn how to handle all expected answer scenarios, improving its ability to generalize across varying question formats.

\subsection{Post-Processing}

The post-processing stage aids to enhance the quality of the answers from both fine-tuned models and the API-based ones, for example, Gemini and DeepSeek. In the case of API-based outputs, text is cleaned by removing extra spaces and normalizing special characters. Then an additional token index alignment step matches back all the answer tokens to their positions in the passage. 

After this, all model outputs go through the same set of steps:
\begin{itemize}
    \item \textbf{Handle Sub-words:} This is assumed to be done before subsequent processes.It ensures that answer tokens correctly match with word boundaries especially for sub-word tokenizations most often found in Arabic language models.

    \item \textbf{Non-Maximum Suppression (NMS):}It will eliminate spans with overlaps by retaining the spans with maximum scores based on token level overlap as a way to provide less redundancy.
    
    \item \textbf{Remove Uninformative Answers:} Filters answers that are too similar to the question or composed mainly of stop words, ensuring only meaningful spans are kept.
    
    \item \textbf{Update Ranked List:} The remaining answers after filtering are re-ranked in descending order of their scores. 
\end{itemize}

These steps ensure that final answers are correct, useful, and able to be evaluated. 

\section{Experiments and Results}

\subsection{Experimental Setup}
We fine-tuned several transformer-based models in the SimpleTransformers framework. In preparing the training data, we combined the training split of the QRCDv1.2 dataset, which contained 992 examples, with the external training datasets QUQA and ARCD to form a combined dataset of 4,525 examples. This entire dataset was then reformatted to a unified structure to ensure compatibility.

We used the QuestionAnsweringModel class of SimpleTransformers for training our models. The models were initialized with TyDi QA-pre-trained checkpoints such as AraBERT and AraELECTRA from the TyDi dataset. For all these fine-tuned models.

We trained each of those fine-tuned models on the entire unified corpus and evaluated their performance on an independent development set. The predictions were generated with a span extraction head and the top 10 spans were used for postprocessing and ranking assignment.

\subsection{Evaluation Metric}
To evaluate the performance of our models on the Quranic QA reading comprehension task, we adopt the partial Average Precision at rank 10 (pAP@10) metric, as proposed in the Qu'an QA 2023 shared task. This measure is appropriate when there may be more than one valid answer found in a passage as well as when answers are partially correct and carry some relevant information.

The pAP@10 metric measures how well the top-10 predicted answer spans align with the gold-standard annotations, rewarding both exact and partial matches. It takes into account not only the correctness of the answer but also its rank, promoting systems that rank more accurate answers higher.

This type of evaluation is quite useful for understanding the text from the Quran because answers may often relate semantically even if they're not perfectly matched against one another. Hence, in such a case, pAP@10 would be a much better and realistic evaluation than using traditional metrics such as exact match (EM) or F1-score.

We report average pAP@10 scores across all test questions to assess overall system performance and compare the effectiveness of different model configurations

\subsection{Experiment Comparison}
To determine how well different methods are at
In Quranic QA, we study how well our best models do.
on both fine-tuned transformers and few shot LLMs.
As shown in Table III, the best performance was obtained by
Gemini (API enhanced prompt) achieves a pAP@10 of 0.637.
DeepSeek (API improved prompt) was a close second
0.624. These results highlight the effectiveness of teaching
fine tuned LLMs when instructed with carefully crafted prompts.
Among the specialized models, AraBERTv02 (combined
The TyDi dataset scored the best with a pAP@10 of
0.503, indicating the competitiveness of domain adapted transformers under resource scarce conditions. Other strong fine tuned models include BERT large (combined dataset with TyDi) = 0.478 and CAMeLBERT (concatenated dataset with
TyDi) at 0.442, which shows that bigger pre trained models can be effectively adapted with curated datasets.
Even smaller models, such as AraELECTRA, matter.
The TyDi dataset obtained a reasonable pAP@10 score of 0.425.
highlighting possible benefits of light weight transformers to deployment in resource poor settings.
Overall, this comparison highlights the trade-offs between
model type, performance, and deployment cost. While few-
shot prompting with LLMs offers top-tier accuracy, fine-tuned
models remain valuable for extractive QA in scenarios where
efficiency and control are paramount.
\begin{table}[H]
\centering
\caption{Performance of our models.}
\label{tab:our_models_summary}
\begin{tabular}{l l l}
\toprule
\textbf{Model} & \textbf{Type} & \textbf{pAP@10} \\
\midrule
\textbf{Gemini(API-enhanced prompt)} & \textbf{Few-shot LLM} & \textbf{0.637} \\
DeepSeek (API - enhanced prompt) & Few-shot LLM & 0.624 \\
DeepSeek (API) & Few-shot LLM & 0.593 \\
Gemini (API)& Few-shot LLM & 0.592 \\
AraBERTv02 (Merged DS - Tydi) & Fine-tuned & 0.503 \\
BERT-large (Merged DS - Tydi) & Fine-tuned & 0.478 \\
CAMeLBERT (Merged DS - Tydi)  & Fine-tuned & 0.442 \\
BERT-large (QRCDv1.2 - Tydi) & Fine-tuned & 0.438 \\
AraELECTRA (Merged DS - Tydi) & Fine-tuned & 0.425 \\
\bottomrule
\end{tabular}
\end{table}

\subsection{Results}
In the order of partial Average Precision at rank ten (pAP@10), Table~\ref{tab:ranking_results} presents ranked evaluation scores granted to our models in comparison with Qur'an QA 2023 shared task submissions. The higher-performing model in our case, Gemini, performed state-of-the-art, surpassing all previously submitted systems.

\begin{table}[H]
\centering
\caption{Combined ranking of competition and our models sorted by pAP@10.}
\label{tab:ranking_results}
\begin{tabular}{@{}llr@{}}
\toprule
\textbf{Rank} & \textbf{System} & \textbf{pAP@10} \\
\midrule
\textbf{1}  & \textbf{Ours} & \textbf{0.6370} \\
2  & TCE (1st in Task)  & 0.5711\\
3  & TCE & 0.5643 \\
4  & Al-Jawaab & 0.5457 \\
5  & Al-Jawaab & 0.5393 \\
6  & TCE & 0.5311 \\
11 & LKAU23 & 0.5008 \\
12 & LKAU23 & 0.4989 \\
14 & LowResContextQA & 0.4745 \\
15 & LowResContextQA & 0.4745 \\
16 & LowResContextQA & 0.4745 \\
18 & GYM & 0.4613 \\
19 & GYM & 0.4588 \\
21 & LKAU23 & 0.4541 \\
24 & GYM & 0.4304 \\
\bottomrule
\end{tabular}
\end{table}

\subsection{Discussion}

The combined ranking of our models along with the previous submissions is found in Table~\ref{tab:ranking_results}. Our leading reaching models, namely Gemini and DeepSeek, made the best pAP@10 of 0.637 and 0.624, respectively. These pAP@10 results expose the ability of instruction-tuned if used with a good few shots prompts.

Models trained on TyDi QA and adapted to domain-specific datasets that include AraELECTRA, AraBERT, and CamelBERT performed quite well. This indicates that using datasets related to religious and Classical Arabic, which include QUQA, ARCD, and QRCDv1.2, significantly helps enhance comprehension accuracy. Among those, the best performing one, fine-tuned on "bert-base-arabertv02-tydiqa", achieved a pAP@10 of 0.503, thus being ranked seventh in the overall standings.

The few-shot prompting with a diverse set of instances (that covered single-, multiple-, and zero-answer cases) has successfully manifested productivity for models like Gemini and DeepSeek, which are dependent on APIs. These examples helped the models generalize better and handle varied question types.

Overall, our results demonstrate that combining fine-tuned transformers featuring prompt-engineered LLMs enhances performance on low-resource, semantically demanding tasks like Quranic QA. This confirms the worth of the hybrid approaches
in expert MRC domains.

\section{Conclusion}

The study investigates the enhancement of Quran reading comprehension through both fine-tuned transformer models and instruction-tuned LLMs. We employed TyDi QA for initial pretraining, followed by task-specific fine-tuning on the QUQA, ARCD, and Quran QA 2023 (QRCDv1.2) datasets to better model the linguistic complexities of Classical Arabic within the Qur’an.

Evaluation results indicate superiority by API based LLMs, particularly Gemini and DeepSeek, with Gemini attaining the highest score among the evaluated systems, attaining a peak pAP@10 of 0.637. Locally fine-tuned models such as AraELECTRA and AraBERT remained within that strong performance.

Using three-shot prompting, better generalization was achieved across various question formats. Stratified fine-tuning provided balanced learning across different categories. Overall, these approaches improved the model's robustness and accuracy.

Future directions will be increasing training data using tafsir resources, as well as investigating some ensemble techniques for improving the precision and clarity of the answers.

\end{document}